\documentclass[letterpaper, 10 pt, conference]{ieee_files/ieeeconf}
\usepackage[utf8]{inputenc}

\usepackage[style=ieee]{biblatex}       
\usepackage{algpseudocode}              
\usepackage{algorithm}                  
\usepackage{graphicx}
\usepackage{subcaption}                 
\usepackage{amsmath}
\usepackage{amssymb}
\usepackage{amsfonts}
\usepackage{hyperref}                   
\usepackage{array}                      
\usepackage{multirow}                   
\graphicspath{ {./figs/} }

\IEEEoverridecommandlockouts                              

\title{\LARGE \bf
Improving Environment Robustness of Deep Reinforcement Learning Approaches for Autonomous Racing Using Bayesian Optimization-based Curriculum Learning
}
\author{Rohan Banerjee$^{1}$*, Prishita Ray$^{1}$*, Mark Campbell$^{2}$
\thanks{*Equal contribution to this work}
\thanks{$^{1}$Computer Science Department, Cornell University  {\tt\small \{rbb242, pr376\}@cornell.edu}}%
\thanks{$^{2}$Mechanical and Aerospace Engineering Department, Cornell University {\tt\small mc288@cornell.edu}}%
}

\begin{document}
\maketitle

\begin{abstract}
    Deep reinforcement learning (RL) approaches have been broadly applied to a large number of robotics tasks, such as robot manipulation and autonomous driving. However, an open problem in deep RL is learning policies that are robust to variations in the environment, which is an important condition for such systems to be deployed into real-world, unstructured settings. Curriculum learning is one approach that has been applied to improve generalization performance in both supervised and reinforcement learning domains, but selecting the appropriate curriculum to achieve robustness can be a user-intensive process. In our work, we show that performing probabilistic inference of the underlying curriculum-reward function using Bayesian Optimization can be a promising technique for finding a robust curriculum. We demonstrate that a curriculum found with Bayesian optimization can outperform a vanilla deep RL agent and a hand-engineered curriculum in the domain of autonomous racing with obstacle avoidance. Our code is available at
    \url{https://github.com/PRISHIta123/Curriculum_RL_for_Driving/}.
\end{abstract}

\section{Introduction}

Deep reinforcement learning approaches have been effective in solving robotic tasks that involve sequential decision-making from high-dimensional observations, such as dexterous manipulation \cite{zeng2018learning} and ground-based navigation \cite{amini2020learning}. For autonomous racing, reinforcement learning approaches have the potential to extend the reach of agents to environments that are unstructured and characterized by a high degree of environmental uncertainty. 

However, one of the open challenges in deep reinforcement learning is that a learned policy may not necessarily generalize well to novel environments that lie outside its training set \cite{cobbe2019quantifying,cobbe2020leveraging,packer2018assessing}. An ideal reinforcement learning agent would be trained in a subset of environments from the distribution of environments that are expected in the real world, and would then be able to generalize to new environments drawn from that same distribution without a significant degradation in performance. For racing, this means that our agent should be able to navigate robustly in the face of differing road configurations (including road topologies), static obstacle configurations, and dynamic obstacles (such as other racing vehicles). 

One set of approaches that have been proposed in the literature to address the generalization problem are approaches based on curriculum learning \cite{bengio2009curriculum}, in which a curated sequence of diverse environments are provided to the learner during the training process. Curriculum-based approaches for RL have been studied in many continuous control domains, including maze environments \cite{dennis2020emergent}, LunarLander \cite{song2022robust}, and autonomous driving \cite{anzalone2021reinforced}, but not extensively for the domain of autonomous racing and obstacle avoidance.

Therefore, the question that we seek to answer in this work is: How can we design deep reinforcement learning-based agents for autonomous racing in the presence of static obstacles, in such a way that they are robust to task-relevant environment variations, such as obstacle and road properties? We propose that curriculum learning can be an effective approach for ensuring robustness in this domain. Specifically, we investigate both manually-designed curricula (based on human domain knowledge), and an automated curriculum generation approach based on applying Bayesian Optimization \cite{frazier2018tutorial} to guide the search through the space of curricula.

We hypothesize that shaping the curriculum in this way can aid with generalization across various reinforcement learning domains with high-dimensional observation spaces, including autonomous racing. Our contributions are as follows:

\begin{itemize}
    \item We show that manually-generated curricula lead to increased robustness to varying obstacle densities and road curvature in a modified version of the OpenAI Gym CarRacing domain with obstacles, compared to policies that are only trained in the default environment parameter setting.
    \item We show that applying Bayesian Optimization (BO) to automatically search for curricula leads to improved robustness performance compared to the manually-generated curriculum in the same domain.
\end{itemize}

\section{Related Work}

A number of papers focus on controlling the scenarios that are encountered during the RL training process to achieve better generalization. This can be done by including as many diverse scenarios as possible in the training set with the expectation that they provide reasonable coverage of the full environment distribution \cite{wang2019autonomous}, or by structuring the training process to include a curriculum of environments \cite{song2022robust},

For autonomous driving, most prior work in the curriculum learning space considers manually-designed curricula, for either street driving scenarios \cite{anzalone2021reinforced} \cite{ozturk2021investigating} \cite{qiao2018automatically} or autonomous overtaking \cite{song2021autonomous}. Other prior work does consider an automatic curriculum generation scheme for driving, either in intersection domains \cite{qiao2018automatically} or racing domains \cite{jiang2021replay} \cite{azad2023clutr}. Our work expands upon the prior work by investigating both manually-designed and automatically-generated curricula (using probabilistic black-box optimization) for the automomous racing domain with static obstacles.

\section{Methodologies}

We examine training curricula that involve changing environment parameters, which we refer to as $\psi \in \Psi$. We represent a curriculum as a list of ordered pairs that represent changepoints in time, at which the agent is exposed to a new environment type $\psi$. Concretely, this means we represent the curriculum $C = [(0, \psi_0), (t_1, \psi_1), \dots, (t_k, \psi_k)]$, where the interpretation is that we expose the learned agent to environments of type $\psi_0$ for epochs in $[0, t_1)$, $\psi_1$ for epochs in $[t_1, t_2)$, and so on.

\subsection{Manual Curriculum Learning}

We first considered manually specifying the curriculum $C$, which for a particular domain and a particular choice of environment parameters $\Psi$ involved manually selecting the epoch sequence $(t_1, \dots, t_k)$ and the environment parameter sequence $(\psi_0, \dots, \psi_k)$. We selected  $(\psi_0, \dots, \psi_k)$ to be an increasing sequence of difficulty (i.e. from easy to difficult), which requires expert domain knowledge about the environment and choice of environment parameters $\Psi$.

\subsection{Automated Curriculum Learning using Bayesian Optimization}

We propose using Bayesian Optimization \cite{frazier2018tutorial} as a method to select the right curriculum. Bayesian optimization starts with a Gaussian Process prior distribution (which is a distribution over functions, in which any subset of input points is multivariate Gaussian), and updates this prior into a posterior distribution over functions, as we collect new data points (using the acquistion function).

The learned function $f: \mathcal{X} \rightarrow \mathbb{R}$ represents a mapping from a particular choice of changepoints to the average expected reward in the ``hard" environment scenarios. The input domain in our case is $\mathcal{X} = \mathbb{R}^k$, where $k$ is the number of epoch changepoints (for the CarRacing domain, we selected $k=3$). Thus, evaluating this function on a particular curriculum $x$ involves training PPO using curriculum $x$ and computing the average expected reward in the ``hard" scenarios. The acquisition function $a: f \times \mathcal{X} \rightarrow \mathbb{R}$ takes in the current Gaussian Process posterior and a possible curriculum $x$, and produces a real-valued score indicating the viability of that particular curriculum.

For the Bayesian Optimization algorithm, we use the following hyperparameters:
\begin{itemize}
    \item Gaussian Process prior mean function $\mu_0: \mathcal{X} \rightarrow \mathbb{R}$: $\mu_0(x) = \mu_0$, where we used $\mu_0 = 0$.
    \item Gaussian Process prior covariance function $\Sigma_0: \mathcal{X} \times \mathcal{X} \rightarrow \mathbb{R}$: we use the Matern kernel with $\nu = 5/2$. We chose this kernel and hyperparameter choice to model functions that are twice differentiable, due to our empirical observation that the reward function was non-smooth but locally quadratic. Additionally, we used feature-dependent length scales $\ell =[19.9,26.5,21.15]$ for the CarRacing domain.
    \item Acquisition function: We use the upper confidence bound (UCB) acquisition function given by $a(x) = \mu(x) + \lambda \sigma(x)$, where $\mu(x)$ and $\sigma(x)$ are the GP posterior mean and standard deviation, respectively. We chose the UCB acquisition function because we empirically observed that the final reward value is highly sensitive to the choice of training curriculum, and thus we desired an exploratory acquisition function. We tried a few values ($\lambda \in [1.75, 1.9, 2]$) and found that $\lambda = 1.9$ worked best empirically.
    \item Optimizer for acquisition function: We select the L-BFGS-B \cite{byrd1995limited} optimization algorithm to minimize $-a(x)$, because it is suited for optimizing nonlinear functions (as is the case for $-a(x)$) and because it is a bounded optimization algorithm, so its search space is more limited, leading to more efficient optimization compared to other methods (e.g. SLSQP). For our domain, we constrained the optimization to lie within the following ranges: $x \in [150-250, 330-450, 730-830]$.
    \item We use $n_0 = 5$ warm-up trials, which were experimentally designed to evenly cover the regions where the local quadratic behavior was empirically observed, and $N = 14$  Bayesian optimization trials.
\end{itemize}

Our automatic curriculum generation algorithm is outlined in Algorithm \ref{alg:bayes-opt}. In practice, we choose the optimal curriculum by not only considering the curriculum $x^*$ with the highest final expected reward $f(x)$, but by also considering all of the the $n_0 + N$ curricula that were discovered by the algorithm. For each trial (curriculum), we examined the entire training reward curve, and we selected the trial that achieved the highest observed reward across all training epochs. For the CarRacing environment, we typically examined only the final 3-4 trials.

\begin{algorithm}
\caption{Bayesian Optimization Curriculum Generator.}
\label{alg:bayes-opt}
\begin{algorithmic}[1]
\State Inputs: Gaussian Process prior hyperparameters (including mean and kernel function), acquisition function $a(\cdot)$, number of warmup points $n_0$, number of iterations $N$
\State Initialize a Gaussian process prior on $f$ and empty dataset $\mathcal{D}$.
\For{$i$ in $[1, n_0]$} \Comment{Warm-Up Phase}
    \State Choose an experimentally-designed curriculum $x_i$.
    \State Observe $y_i = f(x_i)$ by training the PPO agent using curriculum $x_i$.
    \State Update dataset: $\mathcal{D} = \mathcal{D} \cup \{(x_i, y_i)\}$
\EndFor
\For{$n$ in $[1,N]$}
    \State Update the posterior probability distribution on $f$ using $\mathcal{D}$.
    \State Compute $x_n = \arg \max_x a(x)$, the maximizer of the UCB acquisition function over $x$, where the acquisition function is computed using the updated posterior.
    \State Observe $y_i = f(x_i)$ by training the PPO agent using curriculum $x_i$.
    \State Update dataset: $\mathcal{D} = \mathcal{D} \cup \{(x_n, y_n)\}$
\EndFor
\State \Return $x_i$ corresponding to the largest $y_i \in \mathcal{D}$.
\end{algorithmic}
\end{algorithm}

\begin{figure}
    \centering
    \includegraphics[scale=0.225]{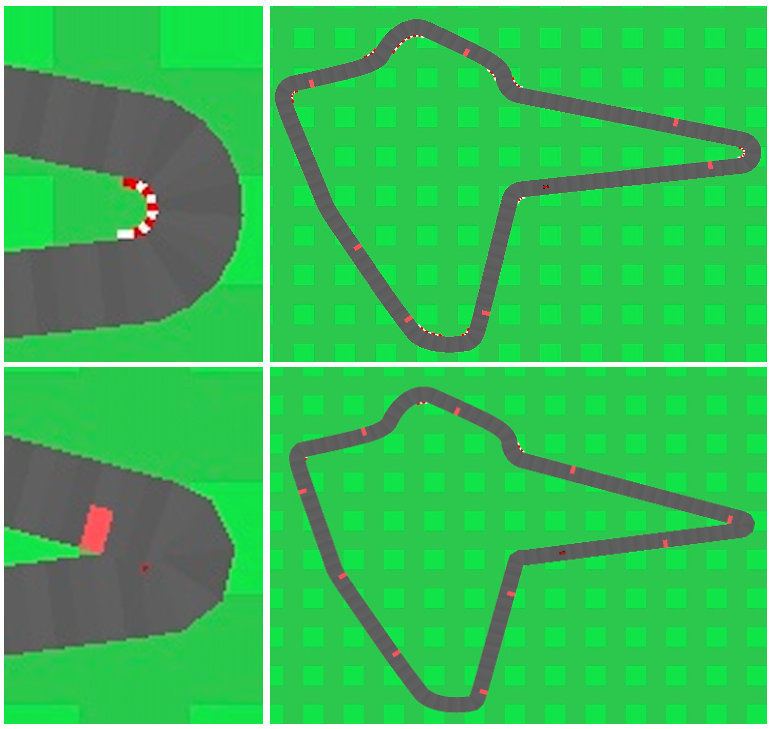}
    \caption{Examples of CarRacing tracks (environment parameters $\psi = [\kappa, p]$ with turn rate $\kappa$ and obstacle probability $p$): (top) $\psi = [0.31, 0.05]$, (bottom) $\psi = [0.71, 0.13]$}
    \label{fig:carracing-images}
\end{figure}

\begin{table*}[h!]   
\caption{Results for three CarRacing curriculum settings. Averages are across 500 evaluation environments.}
\begin{center}
\begin{tabular}{ |c|c|c|c|c|c|c| } 
\hline
\textbf{Training Scheme} & \textbf{Test Setting} &  \textbf{Average Reward} & \textbf{Collision/Obstacle Ratio} & \textbf{Tiles Visited} & \textbf{Time on Grass} & \textbf{Collisions} \\ 
 \hline
 PPO-Default & Easy & $422 \pm 185$ & $0.166$ & $168$ & $0.322$ & 1.002\\ 
  \hline
 PPO-Default & Hard & 
$368 \pm 181$ & $0.185$ & $159$ & $0.343$ & $1.514$ \\ 
 \hline
 Manual Curriculum & Hard & $667 \pm 167$ & $\mathbf{0.055}$ & $230$ & $0.082$ & $\mathbf{0.454}$ \\
 \hline
 BO Curriculum & Hard & $\mathbf{696 \pm 113}$ & $0.133$ & $\mathbf{248}$ & $\mathbf{0.032}$ & $1.088$ \\ 
 \hline

\end{tabular}
\label{table:carracing-1}
\end{center}
\end{table*}

\begin{figure*}[h!]
  \begin{subfigure}{0.5\textwidth}
    \centering
    \includegraphics[scale=0.375]{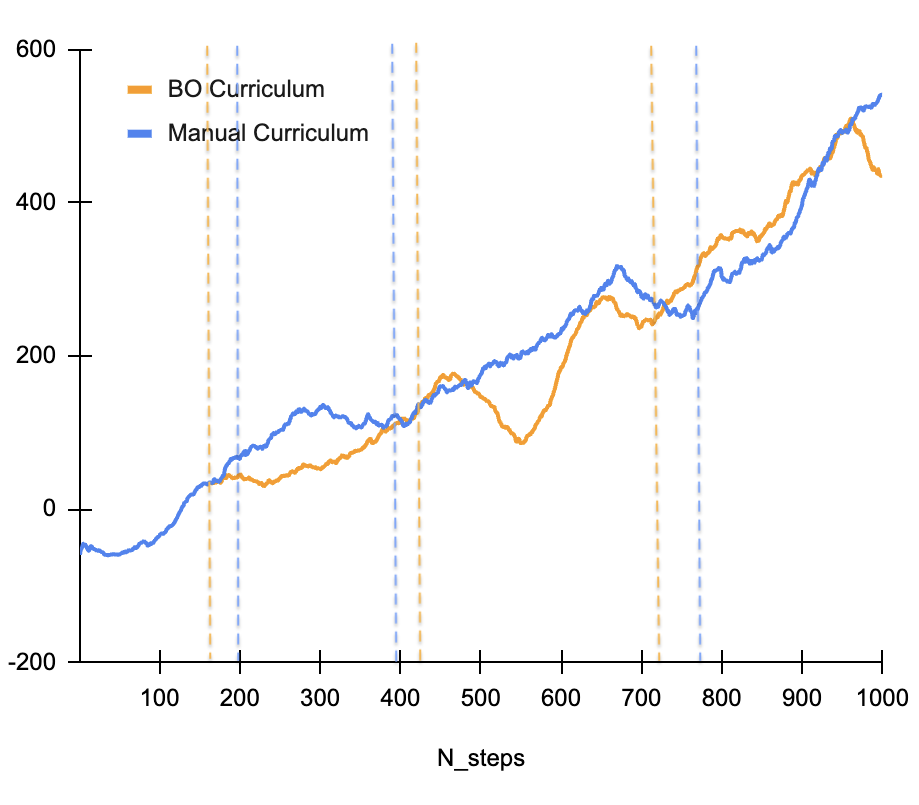}
  \end{subfigure}%
  \begin{subfigure}{0.5\textwidth}
    \centering
    \includegraphics[scale=0.375]{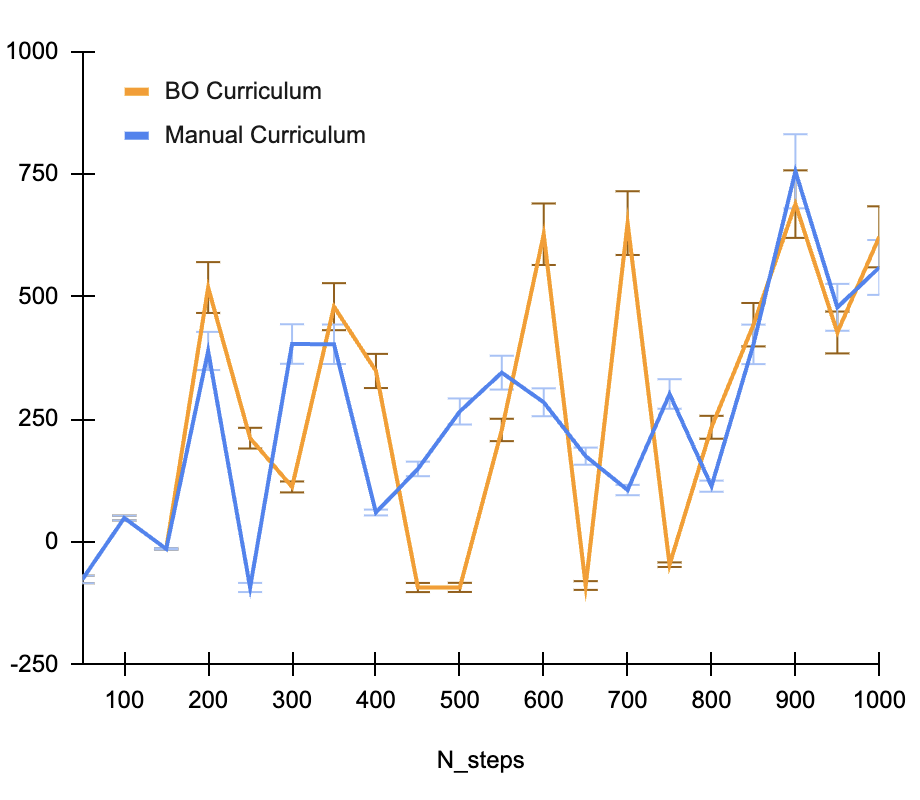}
  \end{subfigure}
  \caption{Mean episode reward comparison between manual curriculum and Bayesian Optimization (BO) curriculum, during (left) training and (right) evaluation. Vertical dashed lines in the training figure indicate curriculum changepoints $t_i$.}
  \label{fig:mean-ep-reward}
\end{figure*}

\begin{figure}[h!]
    \centering
    \includegraphics[scale=0.45]{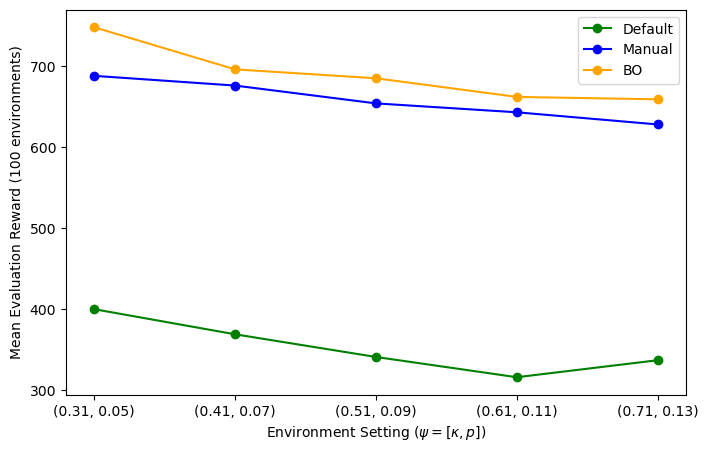}
    \caption{Comparison of evaluation rewards across varying environment difficulty levels for default PPO agent, manual curriculum, and BO curriculum.}
    \label{fig:evaluation-buckets}
\end{figure}

\section{Experimental Setup}
\label{sec:evaluation}

\textbf{Environment.} We evaluate our approach in a modified version of the CarRacing OpenAI Gym environment \cite{carracing}, which is a 2D car navigation environment consisting of a road, static rectangular obstacles, and a grassy area. The objective of the agent in this environment is to complete a lap around the closed-circuit road as quickly as possible while avoiding collisions with the static obstacles. The observation space $\mathcal{O}$ consists of top-down RGB BEV images centered around the car, the action space $\mathcal{A}$ consists of (steering, acceleration, brake) tuples, and the reward function $\mathcal{R}$ is $-0.1$ for every timestep of execution, $-50$ for colliding with obstacles, and $+1000/N_t$ for intersecting a road tile (where $N_t$ is the number of discrete road tiles in the track, typically around 300). Our choice of obstacle penalty means that obstacle collisions have a much larger impact on the reward compared to the number of tiles crossed.

We focus on two key environment variables in our experiments: (1) obstacle density, represented by the probability $p$ that a particular road tile contains a rectangular obstacle, and (2) road curvature, represented by the turn rate $\kappa$ of the road, which dictates the magnitude of turns in the track (that is, a higher $\kappa$ leads to higher curvature turns). In this driving domain, we assume that difficulty increases in proportion to both increased $\kappa$ (turn rate) and $p$ (probability of obstacles). Figure \ref{fig:carracing-images} shows examples of what the driving environment looks like for different choices of $\kappa$ and $p$. The curriculum setting that we considered involved varying both environment variables, so $\psi = [\kappa, p]$. We also considered additional curriculum settings in which we only varied one of the environment variables (either $\psi = [\kappa]$ or $\psi = [p]$), which are described in the Appendix.

\textbf{Base RL algorithm.} We use the PPO algorithm as our underlying deep RL agent \cite{schulman2017proximal}, which we choose due to its simplicity and efficiency as an on-policy deep RL algorithm, and because it has empirically been shown to be more robust than other deep RL methods in navigation domains \cite{larsen2021comparing}. For both the policy and value networks, we use the CNN policy architecture from the Stable Baselines open-source RL library \cite{stable-baselines}.

\textbf{Baselines.} We compare the Bayesian Optimization curriculum learner to two baselines. The first baseline (denoted PPO-Default) involves training PPO on a default set of environment parameters $\psi = [0.31,0.05]$. The second baseline is a manually-selected training curriculum, with an empirically-determined heuristic that we should place more weight on the intermediate-difficulty environments. The manually-chosen curriculum is shown in Table \ref{table:curricula}, indicated with the ranges for each environment instead of the changepoints. We did not consider random curriculum generation as a baseline because empirically it underperformed the manual curriculum.

We train with a single random seed ($0$) for all methods due to our computational constraints. For all methods, we conducted PPO-specific hyperparameter tuning on the default environment using a manual search, and we used the following PPO hyperparameters - learning rate: 0.0002 for curriculum-based training and 0.0005 for default training, number of steps in policy/value update: 10, batch size: 1000, number of training epochs: 1000.

\textbf{Evaluation Methodology.} The robustness evaluation metric we use is the average episode reward on a discrete set $\mathbf{E}_{eval}$, where in our case we consider both a set of easy environments $\mathbf{E}_{easy} = [0.31, 0.05]$ and a set of hard environments $\mathbf{E}_{hard} = h_{\kappa} \times h_{p}$, where $h_{\kappa} = [0.31,0.41,0.51,0.61,0.71]$ and $h_p = [0.05,0.07,0.09,0.11,0.13]$. Given a policy $\pi$ to evaluate, we first sample $N_{eval}$ random environments from $\mathbf{E}_{eval}$, and then compute the mean episodic performance of $\pi$ across the $N_{eval}$ environments. 

\begin{table}[h!]   
\caption{CarRacing Manual and BO training curricula.}
\begin{center}
\begin{tabular}{ |c|c|c| } 
\hline
Manual Epoch Range & BO Epoch Range & $\psi$\\ 
 \hline
0-197 & 0-160 & [0.31, 0.05] \\ 
  \hline
 198-395 & 161-417 & [0.41, 0.07]\\ 
 \hline
 396-774 & 418-736 & [0.51, 0.09] \\
 \hline
 775-1000 & 737-1000 & [0.61, 0.11]\\ 
 \hline
\end{tabular}
\label{table:curricula}
\end{center}
\end{table}

\section{Results and Discussion} 

During model training, we conducted evaluations every $50$ epochs, in which we evaluated the mean and standard deviation of the episodic reward across $N_{eval} = 10$ environments from $\mathbf{E}_{hard}$. Thus Figure \ref{fig:mean-ep-reward} (left) shows the training reward (which indicates the episodic reward of the policy at each epoch in the most-recently generated training environment), and Figure \ref{fig:mean-ep-reward} (right) shows the evaluation reward described earlier, for the manually-selected curriculum baseline and the best curriculum chosen by Bayesian Optimization. Because the training and evaluation scores indicate performance for only 1 and 10 environments, respectively, these estimates tend to have high variance (due to the inherent variability of the environments). Therefore, we decided to evaluate the final learned policies learned from each method across a larger set of environments to produce lower-variance robustness estimates.

In Table \ref{table:carracing-1} we report the mean and standard deviation of the episode return metric (along with other racing-relevant metrics) across $N_{eval}=500$ evaluation environments drawn from $\mathbf{E}_{hard}$ for each algorithm. Even with 19 trials, the Bayesian Optimization method was able to find a curriculum (shown in Table \ref{table:curricula}) that outperformed our baselines. This suggests that the UCB acquisition function was able to find a promising region in curriculum space given the underlying Gaussian Process model of the curriculum reward function. Additionally, Bayesian Optimization was able to manage the tradeoff between getting positive rewards by covering more tiles by staying on the road and a penalty for colliding with obstacles. On an average, CarRacing had $N_t=332$ tiles generated in a track over 500 evaluation environments, and the results in the table conform to the expected behavior.

To further investigate the robustness differences between our methods, we constructed 5 individual sets of $N_{eval} =100$ environments each, ranging from easy (smaller turn rate, lower probability of obstacles) to hard (larger turn rate, higher probability of obstacles). Then, we evaluated the performance of each of the learned policies (PPO-Default, manual curriculum, and Bayesian Optimization) on these 5 sets. Figure \ref{fig:evaluation-buckets} shows the mean evaluation reward for each policy across each environment set. We can observe that the BO Curriculum outperforms the PPO-Default and Manual curriculum baselines on each difficulty level, thus supporting the argument that Bayesian Optimization is able to find better performing curricula from the search space that can generalize well irrespective of difficulty. 

\section{Conclusion}

In this work, we demonstrated that Bayesian Optimization is a potentially promising method to search for robust RL training curricula for autonomous racing. Future work may consider using alternative objective functions for the Bayesian Optimization algorithm, such as the mean of evaluation rewards at different checkpoints during the latter phase of training, or the mean final evaluation reward across multiple random seeds. Either of these alternative objective functions may offer a lower-variance metric for measuring policy robustness to variations in the environment, as performance in a single environment generally has high uncertainty due to the degree of variability in the test environment.

\printbibliography

\newpage

\onecolumn
\appendix

\subsection{Additional Environment Settings}

Here we include results for two different environment settings: $\psi = [\kappa]$ and $\psi = [p]$, which involve varying only the road curvature $\kappa$ or only the obstacle probability $p$, respectively. In the $\psi = [\kappa]$ setting, we set $p = 0$ for all environments, and we define $\mathbf{E}_{easy} = [0.31]$, and $\mathbf{E}_{hard} = h_{\kappa} = [0.31,0.41,0.51,0.61,0.71]$. Similarly, in the $\psi = [p]$ setting, we set $\kappa = 0.31$ for all environments, and we define $\mathbf{E}_{easy} = [0.05]$, and $\mathbf{E}_{hard} = h_{p} = [0.05,0.07,0.09,0.11,0.13]$.

Tables \ref{table:carracing-kappa} and \ref{table:carracing-p} show results analogous to Table \ref{table:carracing-1} for each experimental setting. Note in Table \ref{table:carracing-kappa} we omit metrics related to obstacle collisions because all environments in this setting have $p=0$. In both settings, Bayesian Optimization was able to find curricula that outperformed our baselines, and for the $\psi = [p]$ setting, Bayesian Optimization achieved an improved trade off between positive tile reward and negative obstacle collision reward.

\begin{table*}[h!]   
\caption{Results for three CarRacing curriculum settings with $\psi = [\kappa]$. Averages are across 500 evaluation environments.}
\begin{center}
\begin{tabular}{ |c|c|c|c|c|c|c| } 
\hline
\textbf{Training Scheme} & \textbf{Test Setting} &  \textbf{Average Reward} & \textbf{Tiles Visited} & \textbf{Time on Grass}  \\ 
 \hline
 PPO-Default & Easy & $724 \pm 283$  & $238$ & $0.168$\\ 
  \hline
 PPO-Default & Hard & 
$667 \pm 305$  & $221$ & $0.227$ \\ 
 \hline
 Manual Curriculum & Hard & $682 \pm 121$  & $227$ & $0.157$  \\
 \hline
 BO Curriculum & Hard & $\mathbf{793 \pm 118}$ & $\mathbf{259}$ & $\mathbf{0.042}$ \\ 
 \hline

\end{tabular}
\label{table:carracing-kappa}
\end{center}
\end{table*}

\begin{table*}[h!]   
\caption{Results for three CarRacing curriculum settings with $\psi = [p]$. Averages are across 500 evaluation environments.}
\begin{center}
\begin{tabular}{ |c|c|c|c|c|c|c| } 
\hline
\textbf{Training Scheme} & \textbf{Test Setting} &  \textbf{Average Reward} & \textbf{Collision/Obstacle Ratio} & \textbf{Tiles Visited} & \textbf{Time on Grass} & \textbf{Collisions} \\ 
 \hline
 PPO-Default & Easy & $422 \pm 185$ & $0.166$ & $168$ & $0.322$ & $1.002$\\ 
  \hline
 PPO-Default & Hard & 
$386 \pm 179$ & $0.183$ & $163$ & $0.336$ & $1.418$ \\ 
 \hline
 Manual Curriculum & Hard & $664 \pm 134$ & $0.120$ & $236$ & $\mathbf{0.031}$ & $0.932$ \\
 \hline
 BO Curriculum & Hard & $\mathbf{696 \pm 200}$ & $\mathbf{0.105}$ & $\mathbf{245}$ & $0.125$ & $\mathbf{0.84}$ \\ 
 \hline

\end{tabular}
\label{table:carracing-p}
\end{center}
\end{table*}

\end{document}